# Simplified Minimal Gated Unit Variations for Recurrent Neural Networks

Joel Heck and Fathi M. Salem
Circuits, Systems, and Neural Networks (CSANN) Laboratory
Computer Science and Engineering || Electrical and Computer Engineering
Michigan State University
East Lansing, Michigan 48864-1226
heckjoel@msu.edu; salemf@msu.edu

*Abstract*— Recurrent neural networks with various types of hidden units have been used to solve a diverse range of problems involving sequence data. Two of the most recent proposals, gated recurrent units (GRU) and minimal gated units (MGU), have shown comparable promising results on example public datasets. In this paper, we introduce three model variants of the minimal gated unit (MGU) which further simplify that design by reducing the number of parameters in the forget-gate dynamic equation. These three model variants, referred to simply as MGU1, MGU2, and MGU3, were tested on sequences generated from the MNIST dataset and from the Reuters Newswire Topics (RNT) dataset. The new models have shown similar accuracy to the MGU model while using fewer parameters and thus lowering training expense. One model variant, namely MGU2, performed better than MGU on the datasets considered, and thus may be used as an alternate to MGU or GRU in recurrent neural networks.

## I. INTRODUCTION

Various forms of recurrent units for recurrent neural networks (RNN) have been proposed since the long short-term memory (LSTM) unit in 1997 with impressive results for several sequence-to-sequence applications [1-5]. More recently, simpler recurrent units with fewer gates and fewer parameters have shown comparable performances as these relatively more complicated models [2]. Cho et al. [3] proposed a gated recurrent unit (GRU) in 2014 that uses only two gates and can achieve accuracies comparable to the more complicated LSTM for some applications. Zhou et al. [4] proposed a simpler minimal gated unit (MGU) based on the GRU that only has one gate, namely, the forget gate. In [4], the MGU-based RNN has shown similar accuracy as the GRU-based RNN, but with simpler design and less parameters, and thus less training computational expense.

The recent trend towards simpler recurrent units suggests a need for RNNs with smaller memory footprints and lower training computational load. Previous research has shown that a gated unit would work better than a non-gated simple unit, and so the MGU structure may not be further simplified, as it only has one dynamic gate [4, 5]. However, parameters used in the single gate could justifiably be eliminated to reduce memory footprint and computational expense.

In this paper, we examine and evaluate three simplifications to the MGU model proposed by Zhou et al. [4]. We call these models, MGU1, MGU2, and MGU3, respectively. All three new models and the original MGU model have been comparatively evaluated on standard sequences generated from the MNIST dataset as well as the Reuters Newswire Topics (RNT) dataset. The remainder of the paper is organized as follows: Section II presents a background on the gated recurrent neural networks (RNN), particularly employing the minimal gated unit (MGU), and introduces the proposed designs. Section III specifies the network architectures and Libraries used to evaluate the models. Section IV comparatively summarizes the performance results of all models on both datasets. Finally, Section V concludes the paper and outlines areas for future work.

## II. BACKGROUND

Long Short-Term Memory (LSTM) recurrent neural networks (RNN) have shown impressive results in several applications involving sequence-to-sequence mappings, from speech recognition, translation, to natural language processing [1-3, 5]. They, however, possess relatively complex structure by introducing gated memory units and consequently increase the adaptive parameters by four-fold in comparison to simple recurrent neural networks (sRNN) [1, 2, 5]. Recent research activities have sought to reduce such complexity and reduce the number of parameters to minimize the required memory and computational resources. The gated recurrent units (GRUs) model [2] and the minimal gated units (MGUs) model [4] are examples of such new structures with reduced overall parameters and gating signals. Here, we focus on the minimal gated units (MGUs) model [4] which has been reported to achieve comparable performance to the LSTM and the GRU RNNs in case studies using public datasets [2, 4, 5].

For simple recurrent neural networks (sRNN) [2, 4], the recurrent vector state $h_t$ is updated at each time step according to the following discrete dynamic model:

$$h_t = tanh(Uh_{t-1} + Wx_t + b) \qquad (1)$$

where $x_t$ is the external input vector, tanh is the hyperbolic tangent function, and the parameters are the matrices $W$ and $U$, and vector bias $b$, with appropriate sizes for compatibility.

To be specific, we shall denote the dimensions of the input and recurrent state as m and n, respectively. Then $U$ is an nxn matrix, W is an nxm matrix and b is a nx1 vector.

*A. LSTM RNN:*

It is appropriate to begin the gated RNN modeling evolution from the LSTM RNN which introduced the cell unit structure with its associated gates as follows:

$$c_t = f_t \odot c_{t-1} + i_t \odot \hat{h}_t \qquad (2)$$
$$\hat{h}_t = tanh(Uh_{t-1} + Wx_t + b) \qquad (3)$$
$$h_t = o_t \odot tanh(c_t) \qquad (4)$$

where $c_t$ is referred to as the (vector) memory cell at time t. The LSTM model in Equation (2) incorporates the sRNN model and the previous memory cell value $c_{t-1}$ in an element-wise weighted sum using the forget-gate signal $f_t$ and the input gating signal $i_t$. Note that $\odot$ denotes element-wise (i.e., Hadamard) multiplication. Moreover, in Equation (4), the memory cell is passed through the activation function tanh (.) before (element-wise) multiplying it to the output-gate signal $o_t$ to generate the hidden unit vector $h_t$. Each of the three gate signals is obtained from a replica of the sRNN model using the logistic activation, $\sigma$, instead, to limit its gate signaling range between 0 and 1. Specifically, the gate signals are expressed as:

$$i_t = \sigma(U_i h_{t-1} + W_i x_t + b_i) \qquad (5)$$
$$f_t = \sigma(U_f h_{t-1} + W_f x_t + b_f) \qquad (6)$$
$$o_t = \sigma(U_o h_{t-1} + W_o x_t + b_o) \qquad (7)$$

where each dynamic (vector) gate signal has its own parameters. This constitutes an increase of (adaptive) parameters of approximately four-folds in comparison to the sRNN. We relegate further details to [1, 2, 5].

*B. Gated Recurrent Units (GRU) RNN:*

The gated recurrent units (GRU) [2, 3] simplify the gated RNN down to two gates: an update gate $z_t$ and a reset gate $r_t$. The update gate controls how much the unit updates its j-th activation state $h_t^j$ (in comparison, see Equation (2) above):

$$h_t^j = ((1 - z_t) \odot h_{t-1} + z_t \odot \hat{h}_t)^j \qquad (8)$$

The reset gate controls the amount of history used to update the candidate activation $\hat{h}_t^j$, and would essentially become dependent on only the external signal when it is close to zero. Specifically,

$$\hat{h}_t^j = tanh(U(r_t \odot h_{t-1}) + Wx_t + b)^j \qquad (9)$$

The two gate equations, which are replicas of sRNN using the logistic function $\sigma$, have their own adaptive parameters as expressed in:

$$z_t^j = \sigma(U_z h_{t-1} + W_z x_t + b_z)^j \qquad (10)$$
$$r_t^j = \sigma(U_r h_{t-1} + W_r x_t + b_r)^j \qquad (11)$$

Thus, the GRU RNN increases the adaptive parameters by approximately three-folds in comparison to the sRNN. We relegate further details on the GRU RNN to [2, 3].

*C. Minimal Gated Unit (MGU) RNN:*

The minimal gated unit (MGU) RNN proposed in [4] reduces the number of gates in a GRU from two to one by effectively sharing the update gate with the reset gate. This sharing results in one gate, which is renamed the forget gate *f*, and is computed in the same way as the update gate in Equation (6). Specifically,

$$f_t^j = \sigma(U_f h_{t-1} + W_f x_t + b_f)^j \qquad (12)$$

where the superscript denotes the j-th element of the gate vector. As compared to the GRU RNN, the update equations for the activation state and the candidate activation for the j-th element then become:

$$h_t^j = ((1 - f_t) \odot h_{t-1} + f_t \odot \hat{h}_t)^j \qquad (13)$$
$$\hat{h}_t^j = tanh(U(f_t \odot h_{t-1}) + Wx_t + b)^j \qquad (14)$$

This constitutes an increase of (adaptive) parameters of approximately two-folds in comparison to the sRNN. We relegate further details to [4]. The MGU model has 33% fewer (adaptive) parameters than the GRU model which, as reported in [4], was found to result in faster training over the GRU RNN in the datasets investigated [4].

*1) The MGU RNN Performance*

In [4], Zhou et al. tested an MGU RNN on sequences generated from the MNIST dataset and found comparable or higher accuracies compared to GRU and LSTM RNN. The generated sequences were formed by converting each 28x28 MNIST image, row-wise, to a 28-element vector of 28-length sequences. Analogously, they rolled out the 28x28 image row-wise into a single vector of 784-length sequence. The generated sequences of length 28 or 784 were employed to train 100 hidden MGU RNN over thousands of epochs. Their results for 28-length sequences showed a testing accuracy of 88% for MGU after 16,000 epochs using a batch size of 100. For 784-length sequences, they reported a testing accuracy of 84.25% after 16,000 epochs. These test results were better than GRU under equivalent training and testing conditions.

### D. Simplified Minimal Gated Unit RNN:

In this work, we evaluate three simplifications to the MGU model through variations of the forget gate equation based on control signal considerations and analysis [6]. It is noted that the MGU basic architectural model is expressed by Equations (13)-(14), where the gating signal $f_t$ may be viewed as a control signal. The control signal seeks to achieve the desired sequence-to-sequence mapping using the training data. To that end, its guidance is to minimize the given loss/cost function according to some stopping criterion. This opens up the possibilities of other forms of the control signal besides the one used in Equation (12). This includes the MGU variants considered here due to their simplicity, the reduction in the number of adaptive parameters, and consequently the reduced computational expense. The model variants introduced here are called simply MGU1, MGU2, and MGU3.

#### 1) MGU1

The first variation on the MGU RNN model is to remove the input signal $x_t$ from the gate signal equation (Equation (12)), making the gate dependent only on the unit history and bias:

$$f_t^j = \sigma(U_f h_{t-1} + b_f)^j \quad (15)$$

This variation reduces the number of parameters by the size of the matrix $W_f$, which equals n*m, in comparison to the original MGU RNN model.

#### 2) MGU2

The second variation is to remove the input signal $x_t$ and the bias $b_f$ from the gate signal equation, making the gate dependent only on the unit history:

$$f_t^j = \sigma(U_f h_{t-1})^j \quad (16)$$

This variation further reduces the parameters by the n elements of $b_f$ than the MGU1 model. In total, the reduction equals n*(m+1) in comparison to the original MGU RNN model.

#### 3) MGU3

The third variation is to remove the input signal $x_t$ and the unit state history $h_{t-1}$, leaving just the bias term:

$$f_t^j = \sigma(b_f)^j \quad (17)$$

While Jozefowicz et al. [5] found that the bias term was important in their investigation, here it is unlikely that a gate with just the bias term would result in higher accuracies.

This variation reduces the parameters by n*(n+m) in comparison to the original MGU RNN model. In the case studies in this work, this variation has about 50% of the parameters compared to the (original) MGU model. Thus, the memory foot-print, training and execution would be much faster.

### III. NETWORK ARCHITECTURE

The neural networks for both datasets we tested, MNIST and Reuters Newswire Topics (RNT), were created in Python using the Keras deep learning library [7] and Theano. Since Keras has a GRU layer class, this class was modified to create classes for MGU, the MGU1, MGU2, and MGU3 variations. All of these classes used the hyperbolic tangent function for the candidate activation in Equation (14), and the logistic sigmoid function for the gate activation in Equations (12), (15), (16), and (17).

The MNIST related networks used a batch size of 100 and the RMSProp optimizer [8]. A single layer of hidden units was used with 100 units for the 784-length sequences and 50 units for the 28-length sequences. Although Zhou et al. [4] used 100 units for both lengths of sequences, we decreased the number of units for the shorter sequences to decrease training time. The output layer was a fully connected layer of 10 units in both cases. Table I summarizes the number of (adaptive) parameters used in the MGU, MGU1, MGU2, and MGU3 for the case studies involving units/input dimension/sequence length for the sequences generated from the MNIST dataset.

TABLE I.  NUMBER OF MNIST NETWORK PARAMETERS.

| Units/Input/ Sequence Length | Hidden Unit Type | | | |
|---|---|---|---|---|
| | *MGU* | *MGU1* | *MGU2* | *MGU3* |
| 50 /28 / 28 | 7900 | 6500 | 6450 | 4000 |
| 100 / 1 / 784 | 20400 | 20300 | 20200 | 10300 |

The 28-length sequences were run for 50 epochs, while the 784-length sequences were run for 25 epochs to decrease training time for the longer sequences. Both networks were trained on multiple learning rates for the RMSProp optimizer: $10^{-3}$, $10^{-4}$, and $10^{-5}$.

The RNT dataset was evaluated using a sequence length of 500, 250 units in one hidden layer, and a batch size of 64. The output layer contained 46 fully connected units. Other combinations of sequence length and hidden units were tested, and the best results were with a ratio of 2-to-1. A sequence length of 500 with 250 hidden units was chosen due to time constraints in running the model. Instead of RMSProp, the Adam optimizer [9] was used as it provided slightly better results and ran slightly faster. The learning rate was the default $10^{-3}$ used in [9]. The model was trained across 30 epochs, which was short enough to fit in our time constraints and long enough to show a plateau in the resulting accuracy. Table II summarizes the (adaptive) parameters used in the model variants when using 250 units with sequence dimensions of 500.

TABLE II.  NUMBER OF RNT NETWORK PARAMETERS.

| Units/Input / Sequence Length | Hidden Unit Type | | | |
|---|---|---|---|---|
| | *MGU* | *MGU1* | *MGU2* | *MGU3* |
| 250 /1/ 500 | 126000 | 125750 | 125500 | 63250 |

## IV. Experimental Evaluation

We now summarize the results of the variant MGU models on the MNIST and RNT datasets. The original MGU model results serve as a baseline for comparison to the three variant MGU models.

### A. MNIST dataset

The MNIST database contains 28x28-pixel grayscale images of handwritten digits between zero and nine [10]. These images are separated into a training set of 60,000 images and a test set of 10,000 images. Following procedures in [4, 5] and the references therein, the dataset is evaluated by treating each image as a sequence of 28 elements, each of size (dimension) 28, and alternatively, as a sequence 784 elements, each of size (dimension) one. Using these two varied sequence representations allows for more comparison of the results of the MGU and the variant MGU models. The MNIST database was retrieved from the Keras deep learning library [7].

*1) The 784-Length Generated Sequence*

The best performance on the 784-length MNIST data resulted from a learning rate of $10^{-3}$. Initial performance with that learning rate was inconsistent with significant spikes in the accuracies until the later epochs, as shown in Figure 1. For most of the epochs, MGU2 had the best accuracy, and it achieved slightly better accuracy than MGU after only 25 epochs. The consistent result on this dataset was the relatively poor performance of MGU3 with a learning rate of $10^{-3}$. For a low epoch of 25 and with a learning rate of $10^{-4}$ and $10^{-5}$, MGU3 achieved accuracies similar to the other models, as shown in Table III. This result suggests a need to further verify the performance of the MGU3 model on more case study and longer epoch runs beyond the scope and the sample datasets in this study. It is noted that these studies using the MNIST are referred to as "toy" problems and their values are mainly to establish a baseline comparison among the different variant RNN models.

TABLE III. ACCURACY FOR MNIST-784 AFTER 25 EPOCHS.

| Learning Rate | Hidden Unit Type | | | |
|---|---|---|---|---|
| | *MGU* | *MGU1* | *MGU2* | *MGU3* |
| $10^{-3}$ | 96.8 | 92.8 | 97.1 | 29 |
| $10^{-4}$ | 34.7 | 40.8 | 42.2 | 40.8 |
| $10^{-5}$ | 20.2 | 21.8 | 21.1 | 21.1 |

*2) The 28-Length Generated Sequence*

The performance on the 28-length sequence MNIST data was relatively high after just 50 epochs. Figure 2 shows that the accuracy is above 90% after just several epochs for all models but MGU3 still performs relatively lower. The highest performance resulted from a learning rate of $10^{-3}$, although a rate of $10^{-4}$ was only slightly worse, as shown in Table IV.

Overall, for our hyper-parameter choices, MGU, MGU1, and MGU2 have competitive performances while MGU3 would require perhaps different hyper-parameter settings and/or longer epochs.

TABLE IV. ACCURACY FOR MNIST-28 AFTER 50 EPOCHS.

| Learning Rate | Hidden Unit Type | | | |
|---|---|---|---|---|
| | *MGU* | *MGU1* | *MGU2* | *MGU3* |
| $10^{-3}$ | 97.6 | 98.1 | 98.2 | 96.6 |
| $10^{-4}$ | 95.6 | 95.3 | 94.2 | 91.6 |
| $10^{-5}$ | 69.5 | 71.3 | 71.2 | 65.5 |

For two of the learning rates tested, including the best performance learning rate, MGU1 and MGU2 outperformed MGU by at least 0.5%. MGU3 performed very well even though it only contains the bias term in the gate equation.

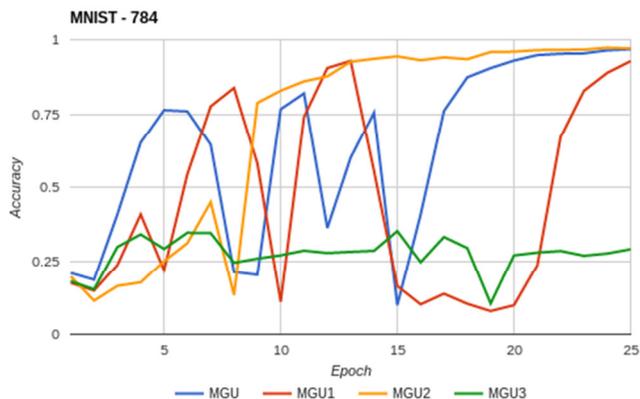

Fig. 1. MNIST 784-length sequence results with a learning rate of $10^{-3}$.

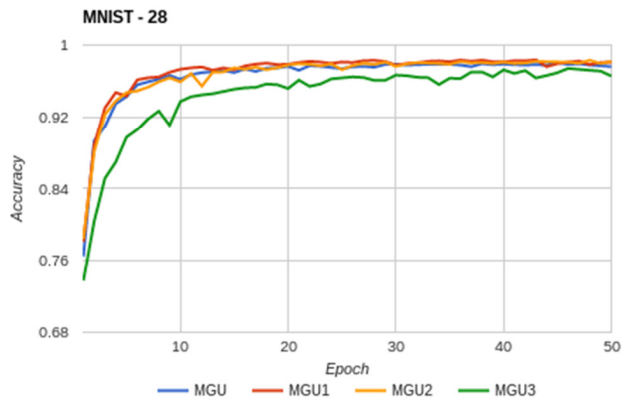

Fig. 2. MNIST 28-length sequence results with a learning rate of $10^{-3}$.

### B. Reuters Newswire Topics (RNT) dataset

The RNT database is a set of 11228 newswire texts collected from the Reuters news agency. These newswires belong to 46 classes based on their topics. The training set includes 8982 newswires, and the testing set includes 2246. Each newswire in the database is preprocessed into a sequence of word indexes with an index corresponding to the overall frequency of a word in the database. This database was retrieved from the Keras deep learning library [7].

TABLE V. ACCURACY FOR RNT AFTER 30 EPOCHS.

| Learning Rate | Hidden Unit Type | | | |
|---|---|---|---|---|
| | *MGU* | *MGU1* | *MGU2* | *MGU3* |
| $10^{-3}$ | 46 | 49.2 | 56.2 | 39.7 |

As with the MNIST database, MGU2 performed the best of the models on the RNT database, improving upon the accuracy of MGU by 22%, as shown in Table V. MGU2 also featured a more consistent accuracy across epochs, unlike the other models which had some notable spikes, as shown in Figure 3. The average per-epoch training time for each model would decrease with fewer parameters.

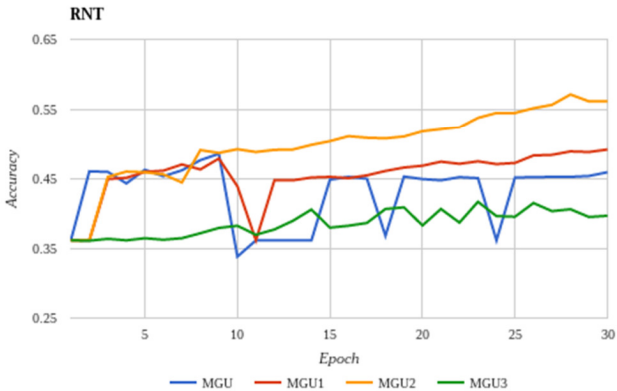

Fig. 3. RNT results with a learning rate of $10^{-3}$.

*C. Discussion*

The MGU variants were tested on two datasets with different types of sequence data (image and text) and different lengths (28, 500, and 784). Compared to MGU, MGU2 provided better accuracy for both datasets. Since MGU2 does not include the input signal nor bias in the gate state equation, it achieved this performance with fewer parameters.

MGU2 also had better accuracy than the other two proposed variants. However, MGU3, which has 50% fewer parameters than MGU, achieved similar, albeit worse, accuracy performance to MGU in two of the tests. This result suggests that a network with MGU3 structures could work reasonably well in certain applications for which a smaller footprint is more critical than achieving the best accuracy.

V. CONCLUSIONS

We described and evaluated three variant models of the original minimal gated unit (MGU) model for use in recurrent neural networks. We simply call these model variants MGU1, MGU2, and MGU3. These three variants were defined by reducing the number of parameters in the forget gate equation in the original MGU. Each of these new model variants has achieved comparable performance to MGU in testing on two popular datasets. The MGU2 variant achieved higher accuracy than the original MGU with fewer parameters and consequently lower training load expense.

Since MGU2 can achieve better performance than MGU, which Zhou et al. [4] showed to achieve comparable performance to GRU, MGU2 could be used in RNN in place of GRU and MGU to achieve similar accuracy and lower training time. Even MGU3 could be used in some cases if fewer parameters and faster training were more important than higher accuracy performance.

Due to resource constraints, the models in this paper could only be tested for low number epochs. A future line of research could run these models for hundreds or thousands of epochs to determine if performance improves, especially for MGU3, and if MGU2 remains better than MGU. It would also be beneficial to run these models on more diverse datasets to gain a better understanding of how the model variants, especially MGU2, compare to MGU across a diverse range of sequence domain applications.